\documentclass[runningheads]{llncs}
\usepackage{graphicx}
\usepackage{wrapfig}
\usepackage{subfig}
\begin{document}
\title{FAIM --  A ConvNet Method for Unsupervised 3D Medical Image Registration}
%
%
\author{Dongyang Kuang\inst{1}
\and
Tanya Schmah\inst{2}}
\authorrunning{D. Kuang et. al}
\institute{University of Ottawa, Ottawa, Canada.\\
	\email{dykuangii@gmail.com} \\
University of Ottawa,Ottawa, Canada. \\
\email{tschmah@uottawa.ca}
}
\maketitle              
\begin{abstract}

We present a new unsupervised learning algorithm, ``FAIM", for 3D medical image registration. With a different architecture than the popular ``U-net"\cite{ronneberger2015u}, the network takes a pair of full image volumes and predicts the displacement fields needed to register source to target. Compared with ``U-net" based registration networks such as VoxelMorph \cite{balakrishnan2018unsupervised}, FAIM has fewer trainable parameters but can achieve higher registration accuracy as
judged by Dice score on region labels in the Mindboggle-101 dataset.  Moreover, with the proposed penalty loss on negative Jacobian determinants, FAIM produces deformations with many fewer ``foldings'', i.e. regions of non-invertibility where the surface folds over itself. In our experiment,  we varied the strength of this penalty and investigated changes in registration accuracy and non-invertibility in terms of number of  ``folding" locations. We found that FAIM is able to maintain both the advantages of higher accuracy and fewer ``folding" locations over VoxelMorph, over a range of hyper-parameters (with the same values used for both networks).
Further, when trading off registration accuracy for better invertibility, FAIM required less sacrifice of registration accuracy. Codes for this paper will be released upon publication.

\end{abstract}
\section{Introduction}

Image registration is a key element of medical image analysis. 
The spatial deformations required to optimally register images are highly non-linear, especially for regions such as the cerebral cortex, 
the folding patterns of which can vary significantly between individuals. 
Most state-of-the-art registration algorithms, such as ANTs \cite{avants2011reproducible},
use geometric or variational methods that are guaranteed to produce diffeomorphisms, i.e. smooth invertible deformations with a smooth inverse. 
These algorithms are very computationally intensive and still do not generally find optimal deformations. 
One general problem is that the optimization problems solved by these algorithms are highly nonconvex.
Another is that they treat each pair of images to be registered \textit{de novo}, without any learning.
\begin{figure}[tbh]
	\centering
	\includegraphics[width=\textwidth]{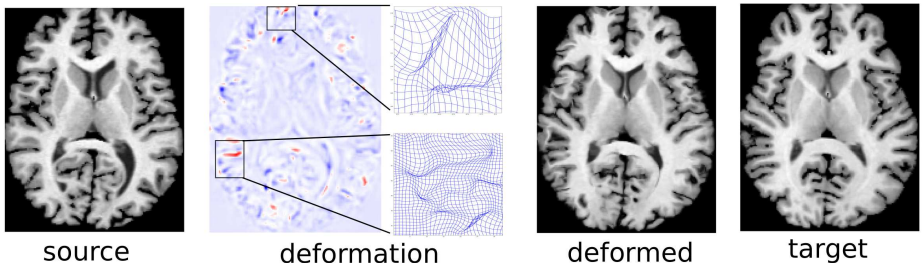}
	\caption{An axial slice of a deformation produced by a CNN method: VoxelMorph-1, with its default $L_2$ regularization parameter $\lambda = 1$ on spatial gradients. The first and last images in the row are the source and target images, while the
	third one is the deformed source image produced by the method.
The second image in the row shows values of the Jacobian determinant of the predicted deformation, with ``folding'' locations (negative determinant) marked in red. The deformed grids illustrate parts of the deformation.}\label{fig:NJ-1}
\end{figure}

A revolution is taking place in the last few years in the application of machine learning methods to medical image processing,
including registration tasks.
Supervised methods for registration, as in \cite{yang2016fast,sokooti2017nonrigid,rohe2017svf}, learn from known reference deformations for training data -- either actual ``ground truth''  in the case of synthetic image pairs, or deformations computed by other automatic or semiautomatic methods.
Unsupervised methods, as in \cite{li2017non,wang2017scalable,shan2017unsupervised,balakrishnan2018unsupervised},
do not require reference deformations, but instead minimize some cost function modeling the goodness of registration, optionally regularized by a term constraining the deformation. These methods have properties complementary to the 
standard geometric methods: they are very fast (at test time) and have the ability to learn automatically from data;
however the predicted deformations are not guaranteed to be diffeomorphisms.
In particular, there are often many regions
where one image has been ``folded'' over itself by a non-invertible transformation.
In these regions the Jacobian matrix of the deformation has negative determinant, as shown in Figure \ref{fig:NJ-1}.
These spatial foldings are not physically possibly and thus constitute registration errors when used in clinical applications.  
The frequency of this kind of error has limited the adoption of neural network methods in medical image registration.

To address this problem, we propose a new unsupervised image registration algorithm, FAIM (for FAst IMage registration) with an explicit anti-folding regularization.
Using the MindBoggle101 dataset \cite{klein2012101}, we compared FAIM's response on both registration accuracy and anti-folding performance with an U-net\cite{ronneberger2015u} based network VoxelMorph \cite{balakrishnan2018unsupervised}. We also examined the trade-off behavior on accuracy and number of foldings on both networks.  

\section{Methods}
\begin{wrapfigure}{L}{8cm}
	\includegraphics[width=.6\textwidth]{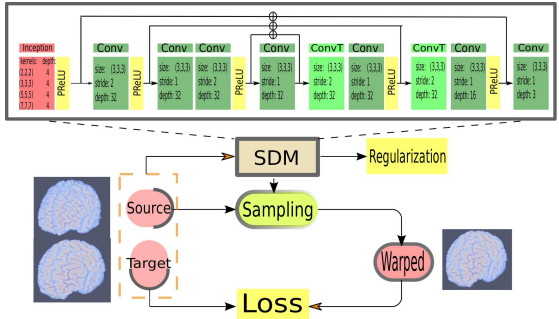}
	\caption{FAIM network architecture. }
	\label{fig:atc}
\end{wrapfigure}
Our architecture is directly inspired by the spatial transformer network (STN) of Jaderberg et al. \cite{jaderberg2015spatial}, which is used to learn the proper parametrized transformation of the input feature so that later tasks such as classifications can be better performed. This kind of module is originally developed for 2D images and only affine and thin plate spline transformation were implemented.  In some very recent research \cite{shan2017unsupervised,balakrishnan2018unsupervised}, this framework begins to appear in 3D medical image registration. 
All these works aim to find an optimal parametrized transformation $\phi: \Omega \rightarrow \mathbf{R}^3$, for image domain $\Omega \subset \mathbf{R}^3$, such that the warped volume $S\circ\phi^{-1} (x)$ from a moving/source volume $S(x)$ is well aligned to the fixed/target volume $T(x)$.
In our network, we use displacement field $\textbf{u}(x)$ to parametrize the deformation $\phi$ by $S\circ \phi^{-1}(x) = S(x+\textbf{u}(x))$, which is learned through a spatial deformation module (SDM). Figure \ref{fig:atc} shows the flow chart when the network is in training and a closer look at the SDM. 

During training, the moving volume and the target volume are stacked together as the input feeding into SDM. The first layer is inspired by Google's Inception module \cite{szegedy2015going}. The purpose of this layer is trying to compare and capture information at different spatial scales for later registration. PReLU \cite{he2015delving} activations are used at the end of each covolutional block except the last layer which uses linear activation to produce displacement fields. The sampling module then takes the displacements and generates a deformed grid and use it to sample the source image to produce the warped image. 
We use kernel stride $>$ 1 to reduce the volume size instead of inserting max pooling layers. Transposed convolutional layers are used for upsampling. There are three ``add" skip connections between the downsampling and upsampling path to help the gradient flow.

The total training loss is 
the sum of an image dissimilarity term $L_{image}$ and regularization terms $L_{total} = L_{image}(S, T) + \alpha R_1({\bf u})+\beta R_2({\bf u})$, 
defined in Table \ref{tab:LossFunc}. 
\begin{table}[tbh]
	\centering
	\begin{tabular}{ll}
		\hline\noalign{\vskip 2mm}
		$L_{image}(S, T)$:  &   $ 1 - CC({S\circ\phi^{-1}}, T)$\\[2mm]  
		Regularization: & $R_1({\bf u}) =  \left\|D \bf u \right\|_{2}$\\[2mm]
		Regularization: & $R_2({\bf u}) = 0.5\, (\, |det (D {\phi^{-1}})| - det (D {\phi^{-1}})\,  )$\\[2mm]
		\hline\noalign{\vskip 2mm}
	\end{tabular}
	\caption{Loss and regularization functions used.}
	\label{tab:LossFunc}
\end{table}
The main loss $L$ with cross correlation (CC) in this paper is for the similarity between the warped source and target, while the first regularization term $R_1$ regularizes the overall smoothness of the predicted displacements. The second regularization aims specifically at penalizing transformations that have many negative Jacobian determinants.
Transformations that have all non-negative Jacobian determinants will not be penalized.

\section{Experiments}
\subsection{Mindboggle101 dataset} This dataset, created by Klein et al. \cite{klein2012101}, is based on a collection of 101 T1-weighted MRIs from healthy subjects. 
The Freesurfer package \\
(\url{http://www.martinos.org/freesurfer})
was used to preprocess all images, and then automatically label the cortex
using its DK cortical parcellation atlas. 
For 54 of the images, including the OASIS-TRT-20 subset, these automatic parcellations were manually edited 
to follow a custom labeling protocol, DKT.  
We use the variant DKT25, with 25 cortical regions per hemisphere.
Details of data collection and processing, including atlas creation,
are described in \cite{klein2012101}.

In the present paper, we used brain volumes from the following three named subsets of Mindboggle101, for a total of 62 volumes: NKI-RS-22, NKI-TRT-20 and OASIS-TRT-20.
These images are already warped to MNI152 space. We normalized the intensity of each brain volume by its maximum voxel intensity. 
Each image 
has dimensions $182 \times 218 \times 182$, which we truncated to $144 \times 180 \times 144$. With this resolution, FAIM has 179,787 trainable parameters, which is about only 70\% of VoxelMorph's 259,675 trainable parameters.

Figure \ref{fig:labels} shows the region corresponding to one label in the parcellation. 
The geometrical complexity of this cortical surface parcellation leads to 
very challenging registration tasks.

\subsection{Evaluation Methods}
We divide each dataset into sets of training and test images, and use these to form training and test sets of \textit{pairs} of images.
The training set consists of all ordered brain volume pairs\footnote{Their corresponding labels are not used in training.} from the union of the NKI-RS-22 and NKI-TRT-20 subsets (1722 pairs in total), and the test set consists of all ordered pairs from the 
OASIS-TRT-20 subset (380 pairs in total).
We train FAIM and VoxelMorph on all pairs of images from the training set,
and then examine their predicted deformations with pairs of images from the test set. 
The Adam optimizer \cite{kingma2014adam} is used. When not otherwise specified, both networks are trained on our training set with the same hyperparameters: learning rate = $10^{-4}$, epochs =10, $\alpha = 1$.


We use predicted deformations to warp corresponding ROI labels from source to target per pair. Registration accuracy is primarily evaluated using the Dice score. 
\begin{equation}
\textbf{Dice}(X, Y) = 2\frac{|X \cap Y|}{|X|+|Y|}.
\end{equation}
It measures the degree of overlap between corresponding regions in the parcellations associated with each image. 
The quality of the predicted deformations $\phi$ is assessed by the total number of locations where Jacobian determinant  $\mathrm{det} \left(\nabla \phi^{-1} (x)\right)$ are negative,
$$\mathcal{N} := \sum \delta(det(D\phi^{-1}) <0).$$
\subsection{Results}\label{sec:results}
\begin{figure}[tbh]
	\centering
	\includegraphics[width=1\textwidth, height=80pt, trim={4cm 6cm 3cm 5cm}, clip]{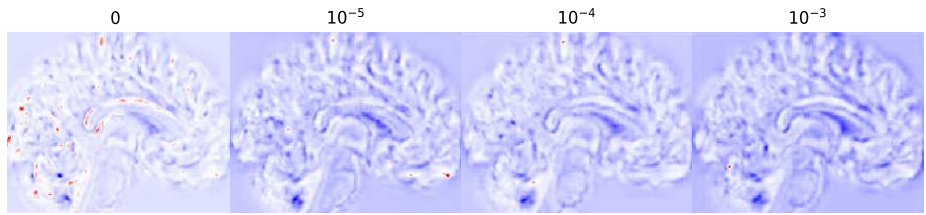}
	\caption{Locations where det$ (D\phi^{-1})<0$ (marked in red) with different $\beta$ shown on one slice. Predictions here are done using FAIM.}
	\label{fig:NJ}
\end{figure}
\begin{figure}[tbh]
	\centering
	\includegraphics[width=.9\textwidth]{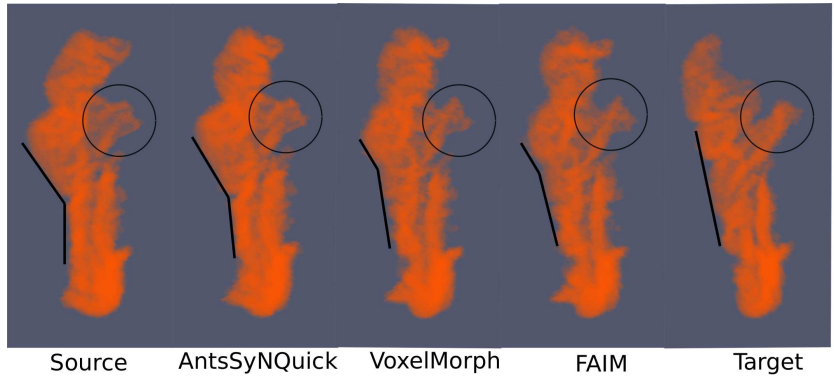}
	\caption{One label (left superior parietal) for one source-target image pair, and the warped source labels produced by different methods. Notice that all three methods are aware of correct regions needed to deform such as thinning the part marked by the circled and straightening the region indicated by the black lines.}
	\label{fig:labels}
\end{figure}

Figure \ref{fig:NJ} visualizes the effect of the second regularization term $R_2(\bf u)$ that penalizes ``foldings" directly during training. When the regularization is not used, $\beta = 0$, there are multiple locations visible in the transformation whose Jacobian determinant are negative. The number is greatly reduced with $\beta=10^{-5}$, and almost eliminated at higher $\beta$ values. Numerical results are given in Table \ref{tab:main}. Figure \ref{fig:labels} provides a visualization of one predicted label rendered in 3D. AntsSyNQuick, Voxelmorph and FAIM appear to produce quite similar results on this label, but the underlying transformations are different as shown in later detailed comparisons. 

\begin{figure}[tbh]
	\centering
\subfloat[Mean Dice scores on separate regions.]{\includegraphics[width=0.49\textwidth]{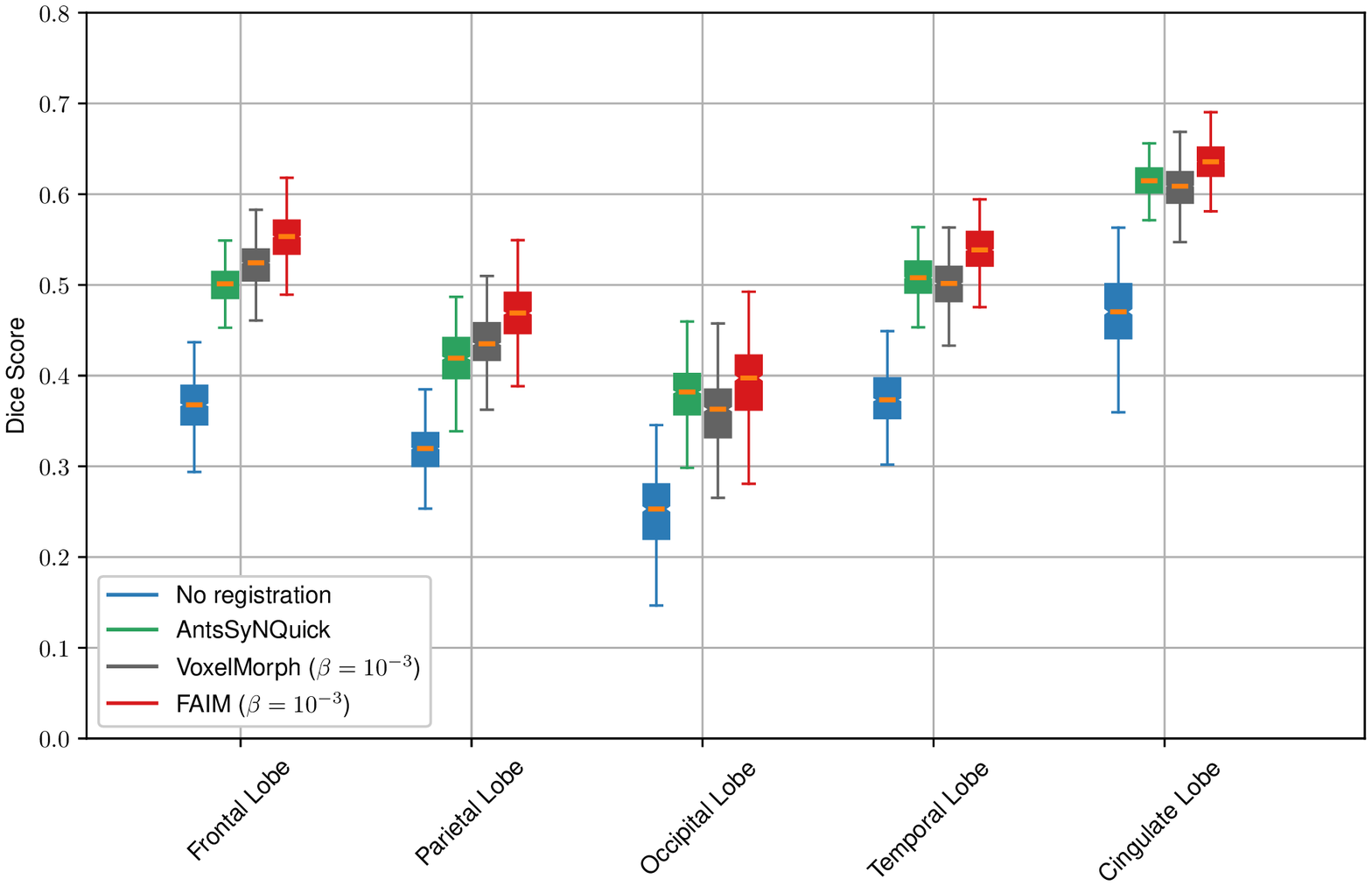}}
\subfloat[Mean Dice score v.s. $\overline{\mathcal{N}}$]{\includegraphics[width=0.49\textwidth]{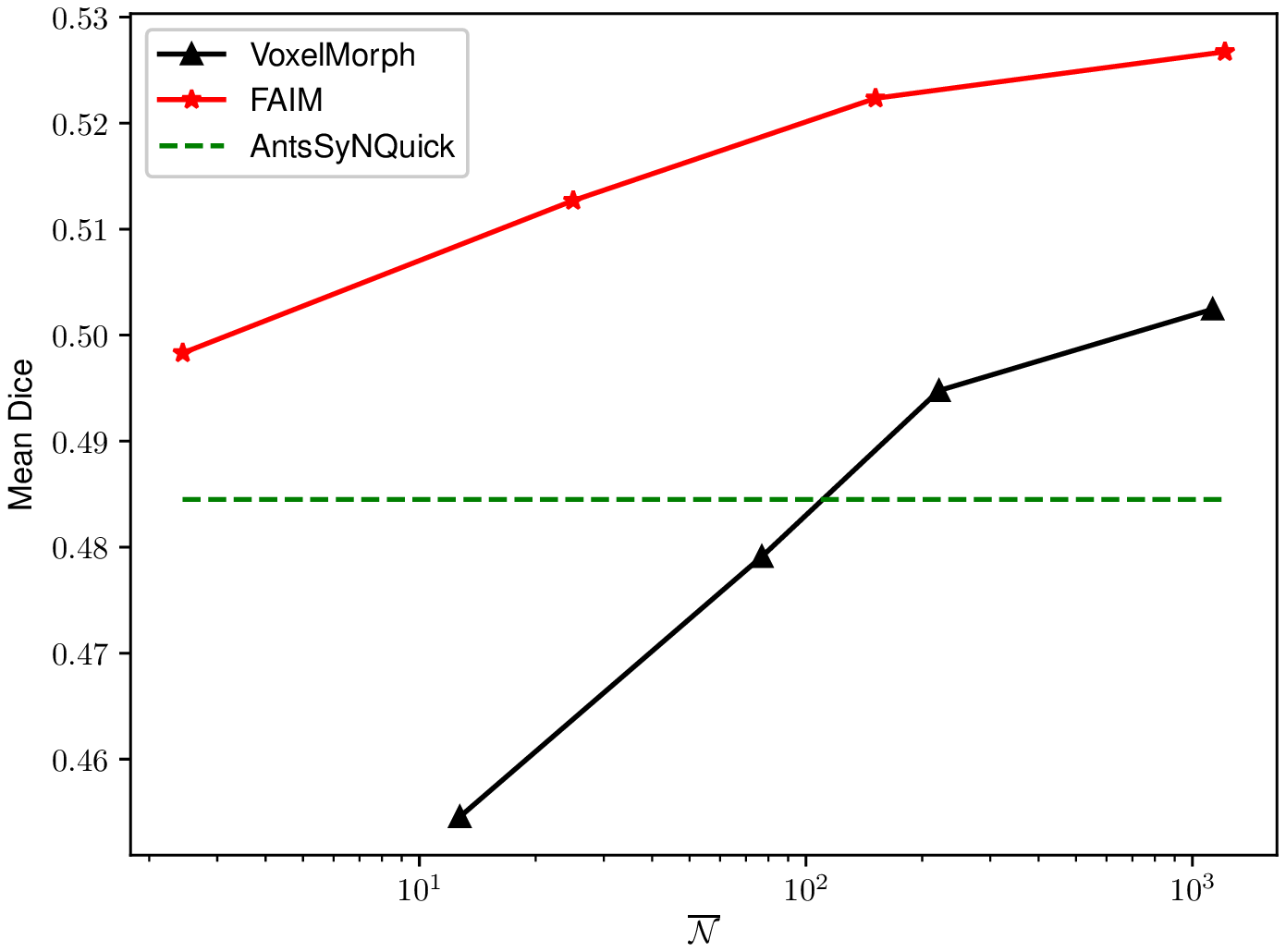}}\\
\subfloat[Mean Dice v.s. $\beta$.]{\includegraphics[width=0.49\textwidth]{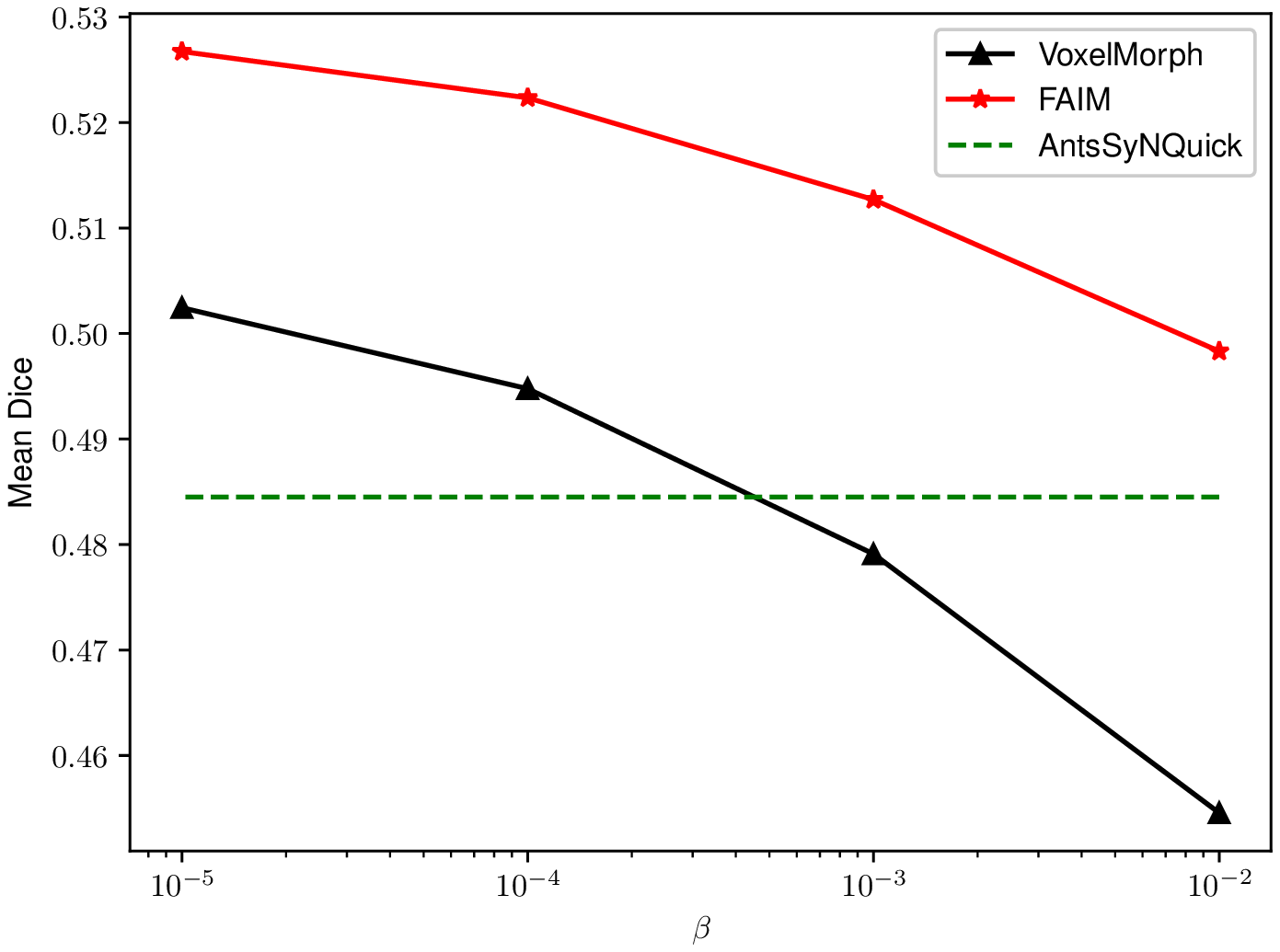}}
\subfloat[ $\overline{\mathcal{N}}$ v.s. $\beta$.]{\includegraphics[width=0.49\textwidth]{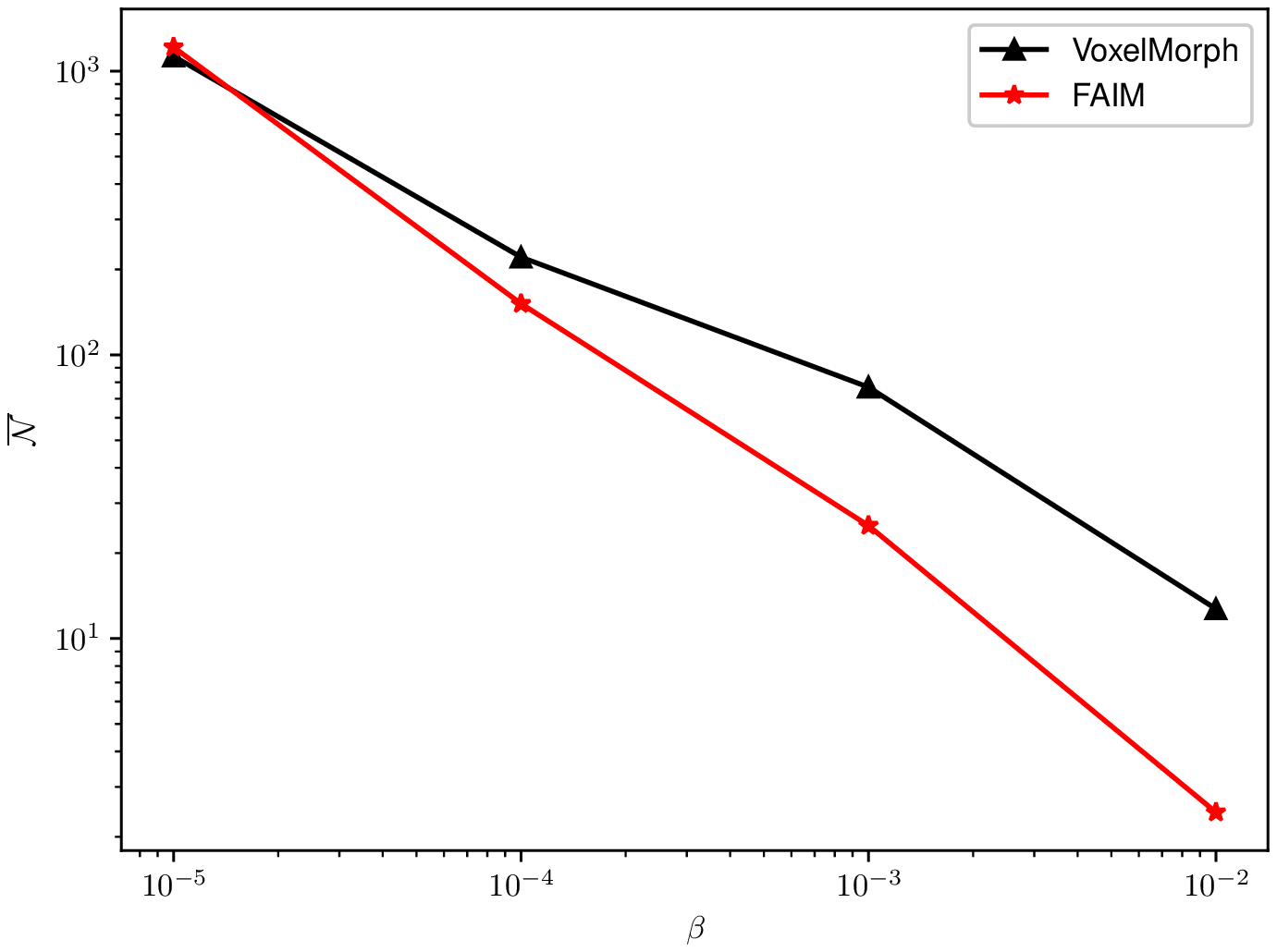}}

\caption{A summary plots of our experiments. In (b) and (c), mean Dice score of AntsSyNQuick are also plotted as a horizontal dashed line free of the two parameters.}
\label{fig:main}
\end{figure}

We selected 5 scales of regularization strength $\beta$ from 0 to $10^{-2}$ and trained both FAIM and VoxelMorph under the same hyper-parameters. We summarize the mean Dice score across all predicted ROI labels with their corresponding target labels in the test set and mean $\mathcal{N}$ (i.e. $\overline{\mathcal{N}}$) of all predicted deformations in the test set in Table \ref{tab:main}.
\begin{table}[tbh]
	\centering
	\begin{tabular}{l|ccccc}
		\hline
		\noalign{\vskip 1mm}
		Mean Dice & $\beta = 0$ &  ${10^{-5}}$ & ${10^{-4}}$ & $10^{-3}$ & $10^{-2}$\\
		\hline
		VoxelMorph &  0.5066&0.5024 & 0.4948& 0.4791& 0.4545\\
		FAIM & \textbf{0.5330} & \textbf{0.5267} & \textbf{0.5230}& \textbf{0.5126}& \textbf{0.4983}\\
		\hline
		\hline
		\noalign{\vskip 1mm}
		Mean $\mathcal{N}$ & $\beta=0$ &  ${10^{-5}}$ & ${10^{-4}}$ & $10^{-3}$ & $10^{-2}$\\
		\hline
		VoxelMorph &  49406& 1129&221 & 77& 13\\
		FAIM &  59115 & 1215& \textbf{151}& \textbf{25}&\textbf{2}\\
		\hline	\noalign{\vskip 1mm}
	\end{tabular}
	\caption{Mean Dice scores and mean number of ``folding" locations with different $\beta $ values. 
	For comparison, the mean Dice score for ANTs SyNQuick is 0.4845.}\label{tab:main}
\end{table}
As one can see in the table, FAIM has higher registration accuracy under all the considered $\beta$ values and lower number of ``foldings" in the predicted deformations when increasing $\beta$. A more detailed comparison on accuracy in terms of Dice score with $\beta = 10^{-3}$ and relations among Dice score, $\overline{\mathcal{N}}$ and $\beta$ are listed together in Figure \ref{fig:main}.

From Figure \ref{fig:main} (a), FAIM has higher registration accuracy among the three compared methods in all the five regions on the brain. Figure \ref{fig:main} (c) suggests this advantage of FAIM in accuracy is consistent across different values of $\beta$, 
with a mean improvement of approximately 3\%. 
To investigate how the networks balance the two competing tasks of high accuracy and low number of  ``folding" locations, we
plotted mean Dice score against mean number of ``folding" locations in Figure \ref{fig:main} (b). In this figure, the flatter curve from FAIM suggests the accuracy of it is more robust with respect to numbers of ``folding" locations in its predictions when compared with VoxelMorph. In other words, the higher slope for VoxelMorph shows that to achieve the same gain in reducing number of ``foldings", U-net based VoxelMorph has to sacrifice more in registration accuracy. Finally, we check the sensitivity of the control of $\beta$ over negative Jacobian determinants in Figure \ref{fig:main} (d) by visualizing $\overline{\mathcal{N}}$ against $\beta$. The sharper slope of FAIM in this log-log plot reveals that we will have more gain in reducing negative Jacobian determinant per unit increase of the regularization strength $\beta$ when compared with VoxelMorph.

\section{Discussion}
We have developed an unsupervised learning algorithm, FAIM, for 3D medical image registration with an option to directly penalize ``foldings", which are spatial locations where the deformation is non-invertible, indicated by a negative 
determinant of the Jacobian matrix. 
Our algorithm is similar to the U-net based registration network VoxelMorph of Balakrishnan et al. \cite{balakrishnan2018unsupervised}, however our architecture design and loss functions are different. 
Our anti-folding penalty is similar to (but different from) the penalty used by Zhang et al. \cite{zhang2018inverse}.
We compared FAIM experimentally to VoxelMorph on the Mindboggle101 dataset \cite{klein2012101}. Our experiments showed that FAIM has advantages in several aspects including: fewer trainable parameters, higher registration accuracy
as measured by Dice score, and less sacrifice needed when trading off registration accuracy for better invertibility 
(fewer ``foldings"). 
In fact, as seen in Table \ref{tab:main}, FAIM with regularization parameter $\beta = 10^{-2}$ produces
deformations that are almost completely invertible (foldings occurring at only 2 voxels per brain on average)
while still having better registration accuracy than the ANTs SyNQuick method.
While we recognise that ANTs is capable of producing more accurate registrations with well-chosen hyperparameters,
our results suggest that NN methods may now be seriously considered for some applications where geometric and variational methods such as ANTs are currently used.

{\small
	\bibliographystyle{splncs04}
	\bibliography{tanya}
}%

\end{document}